\pdfoutput=1

\documentclass[11pt]{article}

\usepackage{EMNLP2023}

\usepackage{times}
\usepackage{latexsym}
\usepackage{listings}

\usepackage[T1]{fontenc}

\usepackage[utf8]{inputenc}

\usepackage{microtype}

\usepackage{inconsolata}

\usepackage{url}

\usepackage{booktabs}
\usepackage{multirow}

\usepackage{amssymb,amsmath}
\usepackage{xcolor}

\title{Ragas: Automated Evaluation of Retrieval Augmented Generation}

\author{Shahul Es$^{\dagger}$, Jithin James$^{\dagger}$, Luis Espinosa-Anke$^{*\diamondsuit}$, Steven Schockaert$^{*}$ \\
        $^{\dagger}$Exploding Gradients \\ 
         $^{*}$CardiffNLP, Cardiff University, United Kingdom \\ 
         $^{\diamondsuit}$AMPLYFI, United Kingdom\\ 
         \texttt{shahules786@gmail.com,jamesjithin97@gmail.com}\\ 
         \texttt{\{espinosa-ankel,schockaerts1\}@cardiff.ac.uk}}

\begin{document}
\maketitle
\begin{abstract}
We introduce \textbf{Ragas} (\textbf{R}etrieval \textbf{A}ugmented \textbf{G}eneration \textbf{As}sessment), a framework for reference-free evaluation of Retrieval Augmented Generation (RAG) pipelines. RAG systems are composed of a retrieval and an LLM based generation module, and provide LLMs with knowledge from a reference textual database, which enables them to act as a natural language layer between a user and textual databases, reducing the risk of hallucinations. Evaluating RAG architectures is, however, challenging because there are several dimensions to consider: the ability of the retrieval system to identify relevant and focused context passages, the ability of the LLM to exploit such passages in a faithful way, or the quality of the generation itself. With Ragas, we put forward a suite of metrics which can be used to evaluate these different dimensions \textit{without having to rely on ground truth human annotations}. We posit that such a framework can crucially contribute to faster evaluation cycles of RAG architectures, which is especially important given the fast adoption of LLMs.

\end{abstract}

\section{Introduction}

Language Models (LMs) capture a vast amount of knowledge about the world, which allows them to answer questions without accessing any external sources. This idea of LMs as repositories of knowledge emerged shortly after the introduction of BERT \cite{devlin-etal-2019-bert} and became more firmly established with the introduction of ever larger LMs \cite{roberts-etal-2020-much}. While the most recent Large Language Models (LLMs) capture enough knowledge to rival human performance across a wide variety of question answering benchmarks \cite{bubeck2023sparks}, the idea of using LLMs as knowledge bases still has two fundamental limitations. First, LLMs are not able to answer questions about events that have happened after they were trained. Second, even the largest models struggle to memorise knowledge that is only rarely mentioned in the training corpus \cite{DBLP:journals/corr/abs-2211-08411,mallen-etal-2023-trust}. The standard solution to these issues is to rely on \emph{Retrieval Augmented Generation (RAG)} \cite{lee2019latent,DBLP:conf/nips/LewisPPPKGKLYR020,guu2020retrieval}. Answering a question then essentially involves retrieving relevant passages from a corpus and feeding these passages, along with the original question, to the LM. While initial approaches relied on specialised LMs for retrieval-augmented language modelling \cite{DBLP:conf/iclr/KhandelwalLJZL20,DBLP:conf/icml/BorgeaudMHCRM0L22}, recent work has suggested that simply adding retrieved documents to the input of a standard LM can also work well \cite{DBLP:journals/corr/abs-2212-14024,DBLP:journals/corr/abs-2302-00083,DBLP:journals/corr/abs-2301-12652}, thus making it possible to use retrieval-augmented strategies in combination with LLMs that are only available through APIs.

While the usefulness of retrieval-augmented strategies is clear, their implementation requires a significant amount of tuning, as the overall performance will be affected by the retrieval model, the considered corpus, the LM, or the prompt formulation, among others. Automated evaluation of retrieval-augmented systems is thus paramount. In practice, RAG systems are often evaluated in terms of the language modelling task itself, i.e.\ by measuring perplexity on some reference corpus. However, such evaluations are not always predictive of downstream performance \cite{DBLP:journals/corr/abs-2305-14625}. Moreover, this evaluation strategy relies on the LM probabilities, which are not accessible for some closed models (e.g.\ ChatGPT and GPT-4). Question answering is another common evaluation task, but usually only datasets with short extractive answers are considered, which may not be representative of how the system will be used.  

To address these issues, in this paper we present Ragas\footnote{Ragas is available at \url{https://github.com/explodinggradients/ragas}.}, a framework for the automated assessment of retrieval augmented generation systems. We focus on settings where reference answers may not be available, and where we want to estimate different proxies for correctness, in addition to the usefulness of the retrieved passages. The Ragas framework provides an integration with both \href{https://github.com/jerryjliu/llama_index}{llama-index} and \href{https://github.com/langchain-ai/langchain}{Langchain}, the most widely used frameworks for building RAG solutions, thus enabling developers to easily integrate Ragas into their standard workflow.

\section{Related Work}

\paragraph{Estimating faithfulness using LLMs}
The problem of detecting hallucinations in LLM generated responses has been extensively studied \cite{ji2023survey}. Several authors have suggested the idea of predicting factuality using a few-shot prompting strategy \cite{DBLP:journals/corr/abs-2304-03728}. Recent analyses, however, suggest that existing models struggle with detecting hallucination when using standard prompting strategies \cite{DBLP:journals/corr/abs-2305-11747,DBLP:journals/corr/abs-2304-13734}. Other approaches rely on linking the generated responses to facts from an external knowledge base \cite{DBLP:journals/corr/abs-2305-14251}, but this is not always possible.

Yet another strategy is to inspect the probabilities assigned to individual tokens, where we would expect the model to be less confident in hallucinated answers than in factual ones. For instance, BARTScore \cite{DBLP:conf/nips/YuanNL21} estimates factuality by looking at the conditional probability of the generated text given the input. \citet{DBLP:journals/corr/abs-2207-05221} use a variation of this idea. Starting from the observation that LLMs provide well-calibrated probabilities when answering multiple-choice questions, they essentially convert the problem of validating model generated answers into a multiple-choice question which asks whether the answer is true or false. Rather than looking at the output probabilities, \citet{DBLP:journals/corr/abs-2304-13734} propose to train a supervised classifier on the weights from one of the hidden layers of the LLM, to predict whether a given statement is true or not. While the approach performs well, the need to access the hidden states of the model makes it unsuitable for systems that access LLMs through an API.

For models that do not provide access to token probabilities, such as ChatGPT and GPT-4, different methods are needed. SelfCheckGPT \cite{DBLP:journals/corr/abs-2303-08896} addresses this problem by instead sampling multiple answers. Their core idea is that factual answers are more stable: when an answer is factual, we can expect that different samples will tend to be semantically similar, whereas this is less likely to be the case for hallucinated answers.

\paragraph{Automated evaluation of text generation systems}
LLMs have also been leveraged to automatically evaluate other aspects of generated text fragments, beyond factuality. For instance, GPTScore \cite{DBLP:journals/corr/abs-2302-04166} uses a prompt that specifies the considered aspect (e.g.\ fluency) and then scores passages based on the average probability of the generated tokens, according to a given autoregressive LM. This idea of using prompts was previously also considered by \citet{DBLP:conf/nips/YuanNL21}, although they used a smaller fine-tuned LM (i.e.\ BART) and did not observe a clear benefit from using prompts. Another approach directly asks ChatGPT to evaluate a particular aspect of the given answer by providing a score between 0 and 100, or by providing a rating on a 5-star scale \cite{DBLP:journals/corr/abs-2303-04048}. Remarkably, strong results can be obtained in this way, although it comes with the limitation of being sensitive to the design of the prompt. Rather than scoring individual answers, some authors have also focused on using an LLM to select the best answer among a number of candidates \cite{DBLP:journals/corr/abs-2305-17926}, typically to compare the performance of different LLMs. However, care is needed with this approach, as the order in which the answers is presented can influence the result \cite{DBLP:journals/corr/abs-2305-17926}. 

In terms of how ground truth answers or, more generally, generations, have been typically used in the literature, most approaches have relied on the availability of one or more reference answers. For instance, BERTScore \cite{DBLP:conf/iclr/ZhangKWWA20} and MoverScore \cite{zhao-etal-2019-moverscore} use contextualised embeddings, produced by a pre-trained BERT model, to compare the similarity between the generated answer and the reference answers. BARTScore \cite{DBLP:conf/nips/YuanNL21} similarly uses reference answers to compute aspects such as precision (estimated as the probability of generating the generated answer given the reference) and recall (estimated as the probability of generating the reference given the generated answer).

\section{Evaluation Strategies}
\label{sec:eval}
We consider a standard RAG setting, where given a question $q$, the system first retrieves some context $c(q)$ and then uses the retrieved context to generate an answer $a_s(q)$. When building a RAG system, we usually do not have access to human-annotated datasets or reference answers. We therefore focus on metrics that are fully self-contained and reference-free. We focus in particular three quality aspects, which we argue are of central importance. First, \textbf{Faithfulness} refers to the idea that the answer should be grounded in the given context. This is important to avoid hallucinations, and to ensure that the retrieved context can act as a justification for the generated answer. 
Indeed, RAG systems are often used in applications where the factual consistency of the generated text w.r.t.\ the grounded sources is highly important, e.g.\ in domains such as law, where information is constantly evolving. Second, \textbf{Answer Relevance} refers to the idea that the generated answer should address the actual question that was provided.  Finally, \textbf{Context Relevance} refers to the idea that the retrieved context should be focused, containing as little irrelevant information as possible. This is important given the cost associated with feeding long context passages to LLMs. Moreover, when context passages are too long, LLMs are often less effective in exploiting that context, especially for information that is provided in the middle of the context passage \cite{liu2023lost}.

We now explain how these three quality aspects can be measured in a fully automated way, by prompting an LLM. In our implementation and experiments, all prompts are evaluated using the \texttt{gpt-3.5-turbo-16k} model, which is available through the OpenAI API\footnote{\url{https://platform.openai.com}}.




\paragraph{Faithfulness}
We say that the answer $a_s(q)$ is faithful to the context $c(q)$ if the claims that are made in the answer can be inferred from the context. To estimate faithfulness, we first use an LLM to extract a set of statements, \( S(a_s(q)) \). The aim of this step is to decompose longer sentences into shorter and more focused assertions. We use the following prompt for this step\footnote{To help clarify the task, we include a demonstration as part of the prompt. This demonstration is not explicitly shown in the listing of the prompts throughout this paper.}:
\begin{quote}
\textit{Given a question and answer, create one or more statements from each sentence in the given answer.\\
question:} \texttt{[question]}\\
\textit{answer:} \texttt{[answer]}
\end{quote}
where \texttt{[question]} and \texttt{[answer]} refer to the given question and answer.
For each statement \( s_i \) in \( S \), the LLM determines if \( s_i \) can be inferred from \( c(q) \) using a verification function \( v(s_i, c(q))\). This verification step is carried out using the following prompt:
\begin{quote}
\textit{Consider the given context and following statements, then determine whether they are supported by the information present in the context. Provide a brief explanation for each statement before arriving at the verdict (Yes/No). Provide a final verdict for each statement in order at the end in the given format. Do not deviate from the specified format.}\\
\textit{statement:} \texttt{[statement 1]}\\
...\\
\textit{statement:} \texttt{[statement $n$]}
\end{quote}
The final faithfulness score, \( F \), is then computed as \( F = \frac{|V|}{|S|} \), where \( |V| \) is the number of statements that were supported according to the LLM and \( |S| \) is the total number of statements.

\paragraph{Answer relevance}
We say that the answer $a_s(q)$ is relevant if it directly addresses the question in an appropriate way. In particular, our assessment of answer relevance does not take into account factuality, but penalises cases where the answer is incomplete or where it contains redundant information. To estimate answer relevance,
for the given answer \( a_s(q) \), we prompt the LLM to generate \( n \) potential questions \( q_i \) based on \( a_s(q) \), as follows:
\begin{quote}
\textit{Generate a question for the given answer.\\
answer}: \texttt{[answer]}
\end{quote}
We then obtain embeddings for all questions using the \texttt{text-embedding-ada-002} model, available from the OpenAI API. For each \( q_i \), we calculate the similarity \( \text{sim}(q, q_i) \) with the original question $q$, as the cosine between the corresponding embeddings.
The answer relevance score, \( \text{AR} \), for question \( q \) is then computed as:

\begin{equation}
\text{AR} = \frac{1}{n} \sum_{i=1}^{n} \text{sim}(q, q_i)
\end{equation}

This metric evaluates how closely the generated answer aligns with the initial question or instruction.

\paragraph{Context relevance}
The context  $c(q)$ is considered relevant to the extent that it exclusively contains information that is needed to answer the question. In particular, this metric aims to penalise the inclusion of redundant information. To estimate context relevance,
given a question \( q \) and its context \( c(q) \), the LLM extracts a subset of sentences, \( S_{ext} \), from \( c(q) \) that are crucial to answer \( q \), using the following prompt:
\begin{quote}
\textit{Please extract relevant sentences from the provided context that can potentially help answer the following question. If no relevant sentences are found, or if you believe the question cannot be answered from the given context, return the phrase "Insufficient Information".  While extracting candidate sentences you're not allowed to make any changes to sentences from given context.}
\end{quote}
The context relevance score is then computed as:
\begin{equation}
\text{CR} = \frac{\text{number of extracted sentences}}{\text{total number of sentences in } c(q)}
\end{equation}






\section{The WikiEval Dataset}
To evaluate the proposed framework, we ideally need examples of question-context-answer triples which are annotated with human judgments. We can then verify to what extent our metrics agree with human assessments of faithfulness, answer relevance and context relevance. Since we are not aware of any publicly available datasets that could be used for this purpose, we created a new dataset, which we refer to as \textit{WikiEval}\footnote{\url{https://huggingface.co/datasets/explodinggradients/WikiEval}}. To construct the dataset, we first selected 50 Wikipedia pages covering events that have happened since the start of 2022\footnote{That is, beyond the reported training cutoff of the model we used in our experiments.}. In selecting these pages, we prioritised those with recent edits. For each of the 50 pages, we then asked ChatGPT to suggest a question that can be answered based on the introductory section of the page, using the following prompt:
\begin{quote}
\textit{Your task is to formulate a question from given context satisfying the rules given below:\\
    1. The question should be fully answered from the given context. \\
    2. The question should be framed from a part that contains non-trivial information.\\ 
    3. The answer should not contain any links.\\ 
    4. The question should be of moderate difficulty.\\
    5. The question must be reasonable and must be understood and responded to by humans.\\
    6. Do not use phrases that 'provided context', etc in the question\\
    context:}
\end{quote}
We also used ChatGPT to answer the generated question, when given the corresponding introductory section as context, using the following prompt:
\begin{quote}
\textit{Answer the question using the information from the given context.\\ 
question:} \texttt{[question]}\\
\textit{context:} \texttt{[context]}
\end{quote} 
All questions were annotated along the three considered quality dimensions by two annotators. Both annotators were fluent in English and were given clear instructions about the meaning of the three considered quality dimensions. For faithfulness and context relevance, the two annotators agreed in around 95\% of cases. For answer relevance, they agreed in around 90\% of the cases. Disagreements were resolved after a discussion between the annotators. 

\paragraph{Faithfulness} To obtain human judgements about faithfulness, we first used ChatGPT to answer the question without access to any additional context. We then asked the annotators to judge which of the two answers was the most faithful (i.e.\ the standard one or the one generated without context), given the question and corresponding Wikipedia page.

\paragraph{Answer relevance}  We first used ChatGPT to obtain candidate answers with lower answer relevance, using the following prompt:
\begin{quote}
\textit{Answer the given question in an incomplete manner.\\
question:} \texttt{[question]}
\end{quote}
We then asked human annotators to compare this answer, and indicate which of the two answers had the highest answer relevance.

\paragraph{Context relevance} To measure this aspect, we first added additional sentences to the context by scraping back-links to the corresponding Wikipedia page. In this way, we were able to add information to the context that was related but less relevant for answering the question. For the few pages without any back-links, we instead used ChatGPT to complete the given context. 


\begin{table}
\begin{tabular}{l@{\hspace{10pt}}c@{\hspace{10pt}}c@{\hspace{10pt}}c}
\toprule
& \textbf{Faith.} & \textbf{Ans.\ Rel.} & \textbf{Cont.\ Rel.}\\
\midrule
Ragas & \textbf{0.95} & \textbf{0.78} & \textbf{0.70} \\
GPT Score & 0.72 & 0.52 & 0.63 \\
GPT Ranking & 0.54 & 0.40 & 0.52 \\
\bottomrule
\end{tabular}
\caption{Agreement with human annotators in pairwise comparisons of faithfulness, answer relevance and context relevance, using the WikEval dataset (accuracy). \label{tabResults}}
\end{table}

\section{Experiments}
\label{sec:experiments}

Table \ref{tabResults} analyses the agreement between the metrics proposed in Section \ref{sec:eval} and the human assessments from the proposed WikiEval dataset. Each WikiEval instance requires the model to compare two answers or two context fragments. We count how often the answer/context preferred by the model (i.e.\ with highest estimated faithfulness, answer relevance, or context relevance) coincides with the answer/context preferred by the human annotators. We report the results in terms of accuracy (i.e.\ the fraction of instances on which the model agrees with the annotators).

To put the results in context, we compare our proposed metrics (shown as \textit{Ragas} in Table \ref{tabResults}) with two baseline methods. For the first method, shown as \emph{GPT Score}, we ask ChatGPT to assign a score between 0 and 10 for the three quality dimensions. To this end, we use a prompt that describes the meaning of the quality metric and then asks to score the given answer/context in line with that definition. For instance, for evaluating faithfulness, we used the following prompt:
\begin{quote}
\textit{Faithfulness measures the information consistency of the answer against the given context. Any claims that are made in the answer that cannot be deduced from context should be penalized.\\
Given an answer and context, assign a score for faithfulness in the range 0-10.\\
context}: \texttt{[context]}\\
\textit{answer}: \texttt{[answer]}
\end{quote}
Ties, where the same score is assigned by the LLM to both answer candidates, were broken randomly.
The second baseline, shown as \emph{GPT Ranking}, instead asks ChatGPT to select the preferred answer/context. In this case, the prompt again includes a definition of the considered quality metric. For instance, for evaluating answer relevance, we used the following prompt:
\begin{quote}
\textit{Answer Relevancy measures the degree to which a response directly addresses and is appropriate for a given question. It penalizes the present of redundant information or incomplete answers given a question.
Given an question and answer, rank each answer based on Answer Relevancy.\\
question}: \texttt{[question]}\\
\textit{answer 1}: \texttt{[answer 1]}\\
\textit{answer 2}: \texttt{[answer 2]}
\end{quote}

The results in Table \ref{tabResults} show that our proposed metrics are much closer aligned with the human judgements than the predictions from the two baselines. For faithfulness, the Ragas prediction are in general highly accurate. For answer relevance, the agreement is lower, but this is largely due to the fact that the differences between the two candidate answers are often very subtle. We found context relevance to be the hardest quality dimension to evaluate. In particular, we observed that ChatGPT often struggles with the task of selecting the sentences from the context that are crucial, especially for longer contexts. 


\section{Conclusions}
We have highlighted the need for automated reference-free evaluation of RAG systems. In particular, we have argued the need for an evaluation framework that can assess faithfulness (i.e.\ is the answer grounded in the retrieved context), answer relevance (i.e.\ does the answer address the question) and context relevance (i.e.\ is the retrieved context sufficiently focused). To support the development of such a framework, we have introduced \textit{WikiEval}, a dataset which human judgements of these three different aspects. Finally, we have also described Ragas, our implementation of the three considered quality aspects. This framework is easy to use and can provide deverlopers of RAG systems with valuable insights, even in the absence of any ground truth. Our evaluation on WikiEval has shown that the predictions from Ragas are closely aligned with human predictions, especially for faithfulness and answer relevance.



\bibliography{anthology,custom}

\begin{thebibliography}{28}
\expandafter\ifx\csname natexlab\endcsname\relax\def\natexlab#1{#1}\fi

\bibitem[{Azaria and Mitchell(2023)}]{DBLP:journals/corr/abs-2304-13734}
Amos Azaria and Tom~M. Mitchell. 2023.
\newblock \href {https://doi.org/10.48550/arXiv.2304.13734} {The internal state of an {LLM} knows when its lying}.
\newblock \emph{CoRR}, abs/2304.13734.

\bibitem[{Borgeaud et~al.(2022)Borgeaud, Mensch, Hoffmann, Cai, Rutherford, Millican, van~den Driessche, Lespiau, Damoc, Clark, de~Las~Casas, Guy, Menick, Ring, Hennigan, Huang, Maggiore, Jones, Cassirer, Brock, Paganini, Irving, Vinyals, Osindero, Simonyan, Rae, Elsen, and Sifre}]{DBLP:conf/icml/BorgeaudMHCRM0L22}
Sebastian Borgeaud, Arthur Mensch, Jordan Hoffmann, Trevor Cai, Eliza Rutherford, Katie Millican, George van~den Driessche, Jean{-}Baptiste Lespiau, Bogdan Damoc, Aidan Clark, Diego de~Las~Casas, Aurelia Guy, Jacob Menick, Roman Ring, Tom Hennigan, Saffron Huang, Loren Maggiore, Chris Jones, Albin Cassirer, Andy Brock, Michela Paganini, Geoffrey Irving, Oriol Vinyals, Simon Osindero, Karen Simonyan, Jack~W. Rae, Erich Elsen, and Laurent Sifre. 2022.
\newblock \href {https://proceedings.mlr.press/v162/borgeaud22a.html} {Improving language models by retrieving from trillions of tokens}.
\newblock In \emph{International Conference on Machine Learning, {ICML} 2022, 17-23 July 2022, Baltimore, Maryland, {USA}}, volume 162 of \emph{Proceedings of Machine Learning Research}, pages 2206--2240. {PMLR}.

\bibitem[{Bubeck et~al.(2023)Bubeck, Chandrasekaran, Eldan, Gehrke, Horvitz, Kamar, Lee, Lee, Li, Lundberg et~al.}]{bubeck2023sparks}
S{\'e}bastien Bubeck, Varun Chandrasekaran, Ronen Eldan, Johannes Gehrke, Eric Horvitz, Ece Kamar, Peter Lee, Yin~Tat Lee, Yuanzhi Li, Scott Lundberg, et~al. 2023.
\newblock Sparks of artificial general intelligence: Early experiments with gpt-4.
\newblock \emph{arXiv preprint arXiv:2303.12712}.

\bibitem[{Devlin et~al.(2019)Devlin, Chang, Lee, and Toutanova}]{devlin-etal-2019-bert}
Jacob Devlin, Ming-Wei Chang, Kenton Lee, and Kristina Toutanova. 2019.
\newblock \href {https://doi.org/10.18653/v1/N19-1423} {{BERT}: Pre-training of deep bidirectional transformers for language understanding}.
\newblock In \emph{Proceedings of the 2019 Conference of the North {A}merican Chapter of the Association for Computational Linguistics: Human Language Technologies, Volume 1 (Long and Short Papers)}, pages 4171--4186, Minneapolis, Minnesota. Association for Computational Linguistics.

\bibitem[{Fu et~al.(2023)Fu, Ng, Jiang, and Liu}]{DBLP:journals/corr/abs-2302-04166}
Jinlan Fu, See{-}Kiong Ng, Zhengbao Jiang, and Pengfei Liu. 2023.
\newblock \href {https://doi.org/10.48550/arXiv.2302.04166} {Gptscore: Evaluate as you desire}.
\newblock \emph{CoRR}, abs/2302.04166.

\bibitem[{Guu et~al.(2020)Guu, Lee, Tung, Pasupat, and Chang}]{guu2020retrieval}
Kelvin Guu, Kenton Lee, Zora Tung, Panupong Pasupat, and Mingwei Chang. 2020.
\newblock Retrieval augmented language model pre-training.
\newblock In \emph{International conference on machine learning}, pages 3929--3938. PMLR.

\bibitem[{Ji et~al.(2023)Ji, Lee, Frieske, Yu, Su, Xu, Ishii, Bang, Madotto, and Fung}]{ji2023survey}
Ziwei Ji, Nayeon Lee, Rita Frieske, Tiezheng Yu, Dan Su, Yan Xu, Etsuko Ishii, Ye~Jin Bang, Andrea Madotto, and Pascale Fung. 2023.
\newblock Survey of hallucination in natural language generation.
\newblock \emph{ACM Computing Surveys}, 55(12):1--38.

\bibitem[{Kadavath et~al.(2022)Kadavath, Conerly, Askell, Henighan, Drain, Perez, Schiefer, Hatfield{-}Dodds, DasSarma, Tran{-}Johnson, Johnston, Showk, Jones, Elhage, Hume, Chen, Bai, Bowman, Fort, Ganguli, Hernandez, Jacobson, Kernion, Kravec, Lovitt, Ndousse, Olsson, Ringer, Amodei, Brown, Clark, Joseph, Mann, McCandlish, Olah, and Kaplan}]{DBLP:journals/corr/abs-2207-05221}
Saurav Kadavath, Tom Conerly, Amanda Askell, Tom Henighan, Dawn Drain, Ethan Perez, Nicholas Schiefer, Zac Hatfield{-}Dodds, Nova DasSarma, Eli Tran{-}Johnson, Scott Johnston, Sheer~El Showk, Andy Jones, Nelson Elhage, Tristan Hume, Anna Chen, Yuntao Bai, Sam Bowman, Stanislav Fort, Deep Ganguli, Danny Hernandez, Josh Jacobson, Jackson Kernion, Shauna Kravec, Liane Lovitt, Kamal Ndousse, Catherine Olsson, Sam Ringer, Dario Amodei, Tom Brown, Jack Clark, Nicholas Joseph, Ben Mann, Sam McCandlish, Chris Olah, and Jared Kaplan. 2022.
\newblock \href {https://doi.org/10.48550/arXiv.2207.05221} {Language models (mostly) know what they know}.
\newblock \emph{CoRR}, abs/2207.05221.

\bibitem[{Kandpal et~al.(2022)Kandpal, Deng, Roberts, Wallace, and Raffel}]{DBLP:journals/corr/abs-2211-08411}
Nikhil Kandpal, Haikang Deng, Adam Roberts, Eric Wallace, and Colin Raffel. 2022.
\newblock \href {https://doi.org/10.48550/arXiv.2211.08411} {Large language models struggle to learn long-tail knowledge}.
\newblock \emph{CoRR}, abs/2211.08411.

\bibitem[{Khandelwal et~al.(2020)Khandelwal, Levy, Jurafsky, Zettlemoyer, and Lewis}]{DBLP:conf/iclr/KhandelwalLJZL20}
Urvashi Khandelwal, Omer Levy, Dan Jurafsky, Luke Zettlemoyer, and Mike Lewis. 2020.
\newblock \href {https://openreview.net/forum?id=HklBjCEKvH} {Generalization through memorization: Nearest neighbor language models}.
\newblock In \emph{8th International Conference on Learning Representations, {ICLR} 2020, Addis Ababa, Ethiopia, April 26-30, 2020}. OpenReview.net.

\bibitem[{Khattab et~al.(2022)Khattab, Santhanam, Li, Hall, Liang, Potts, and Zaharia}]{DBLP:journals/corr/abs-2212-14024}
Omar Khattab, Keshav Santhanam, Xiang~Lisa Li, David Hall, Percy Liang, Christopher Potts, and Matei Zaharia. 2022.
\newblock \href {https://doi.org/10.48550/arXiv.2212.14024} {Demonstrate-search-predict: Composing retrieval and language models for knowledge-intensive {NLP}}.
\newblock \emph{CoRR}, abs/2212.14024.

\bibitem[{Lee et~al.(2019)Lee, Chang, and Toutanova}]{lee2019latent}
Kenton Lee, Ming-Wei Chang, and Kristina Toutanova. 2019.
\newblock Latent retrieval for weakly supervised open domain question answering.
\newblock In \emph{Proceedings of the 57th Annual Meeting of the Association for Computational Linguistics}, pages 6086--6096.

\bibitem[{Lewis et~al.(2020)Lewis, Perez, Piktus, Petroni, Karpukhin, Goyal, K{\"{u}}ttler, Lewis, Yih, Rockt{\"{a}}schel, Riedel, and Kiela}]{DBLP:conf/nips/LewisPPPKGKLYR020}
Patrick S.~H. Lewis, Ethan Perez, Aleksandra Piktus, Fabio Petroni, Vladimir Karpukhin, Naman Goyal, Heinrich K{\"{u}}ttler, Mike Lewis, Wen{-}tau Yih, Tim Rockt{\"{a}}schel, Sebastian Riedel, and Douwe Kiela. 2020.
\newblock \href {https://proceedings.neurips.cc/paper/2020/hash/6b493230205f780e1bc26945df7481e5-Abstract.html} {Retrieval-augmented generation for knowledge-intensive {NLP} tasks}.
\newblock In \emph{Advances in Neural Information Processing Systems 33: Annual Conference on Neural Information Processing Systems 2020, NeurIPS 2020, December 6-12, 2020, virtual}.

\bibitem[{Li et~al.(2023)Li, Cheng, Zhao, Nie, and Wen}]{DBLP:journals/corr/abs-2305-11747}
Junyi Li, Xiaoxue Cheng, Wayne~Xin Zhao, Jian{-}Yun Nie, and Ji{-}Rong Wen. 2023.
\newblock \href {https://doi.org/10.48550/arXiv.2305.11747} {Halueval: {A} large-scale hallucination evaluation benchmark for large language models}.
\newblock \emph{CoRR}, abs/2305.11747.

\bibitem[{Liu et~al.(2023)Liu, Lin, Hewitt, Paranjape, Bevilacqua, Petroni, and Liang}]{liu2023lost}
Nelson~F. Liu, Kevin Lin, John Hewitt, Ashwin Paranjape, Michele Bevilacqua, Fabio Petroni, and Percy Liang. 2023.
\newblock \href {http://arxiv.org/abs/2307.03172} {Lost in the middle: How language models use long contexts}.

\bibitem[{Mallen et~al.(2023)Mallen, Asai, Zhong, Das, Khashabi, and Hajishirzi}]{mallen-etal-2023-trust}
Alex Mallen, Akari Asai, Victor Zhong, Rajarshi Das, Daniel Khashabi, and Hannaneh Hajishirzi. 2023.
\newblock \href {https://aclanthology.org/2023.acl-long.546} {When not to trust language models: Investigating effectiveness of parametric and non-parametric memories}.
\newblock In \emph{Proceedings of the 61st Annual Meeting of the Association for Computational Linguistics (Volume 1: Long Papers)}, pages 9802--9822, Toronto, Canada. Association for Computational Linguistics.

\bibitem[{Manakul et~al.(2023)Manakul, Liusie, and Gales}]{DBLP:journals/corr/abs-2303-08896}
Potsawee Manakul, Adian Liusie, and Mark J.~F. Gales. 2023.
\newblock \href {https://doi.org/10.48550/arXiv.2303.08896} {Selfcheckgpt: Zero-resource black-box hallucination detection for generative large language models}.
\newblock \emph{CoRR}, abs/2303.08896.

\bibitem[{Min et~al.(2023)Min, Krishna, Lyu, Lewis, Yih, Koh, Iyyer, Zettlemoyer, and Hajishirzi}]{DBLP:journals/corr/abs-2305-14251}
Sewon Min, Kalpesh Krishna, Xinxi Lyu, Mike Lewis, Wen{-}tau Yih, Pang~Wei Koh, Mohit Iyyer, Luke Zettlemoyer, and Hannaneh Hajishirzi. 2023.
\newblock \href {https://doi.org/10.48550/arXiv.2305.14251} {Factscore: Fine-grained atomic evaluation of factual precision in long form text generation}.
\newblock \emph{CoRR}, abs/2305.14251.

\bibitem[{Ram et~al.(2023)Ram, Levine, Dalmedigos, Muhlgay, Shashua, Leyton{-}Brown, and Shoham}]{DBLP:journals/corr/abs-2302-00083}
Ori Ram, Yoav Levine, Itay Dalmedigos, Dor Muhlgay, Amnon Shashua, Kevin Leyton{-}Brown, and Yoav Shoham. 2023.
\newblock \href {https://doi.org/10.48550/arXiv.2302.00083} {In-context retrieval-augmented language models}.
\newblock \emph{CoRR}, abs/2302.00083.

\bibitem[{Roberts et~al.(2020)Roberts, Raffel, and Shazeer}]{roberts-etal-2020-much}
Adam Roberts, Colin Raffel, and Noam Shazeer. 2020.
\newblock \href {https://doi.org/10.18653/v1/2020.emnlp-main.437} {How much knowledge can you pack into the parameters of a language model?}
\newblock In \emph{Proceedings of the 2020 Conference on Empirical Methods in Natural Language Processing (EMNLP)}, pages 5418--5426, Online. Association for Computational Linguistics.

\bibitem[{Shi et~al.(2023)Shi, Min, Yasunaga, Seo, James, Lewis, Zettlemoyer, and Yih}]{DBLP:journals/corr/abs-2301-12652}
Weijia Shi, Sewon Min, Michihiro Yasunaga, Minjoon Seo, Rich James, Mike Lewis, Luke Zettlemoyer, and Wen{-}tau Yih. 2023.
\newblock \href {https://doi.org/10.48550/arXiv.2301.12652} {{REPLUG:} retrieval-augmented black-box language models}.
\newblock \emph{CoRR}, abs/2301.12652.

\bibitem[{Wang et~al.(2023{\natexlab{a}})Wang, Liang, Meng, Shi, Li, Xu, Qu, and Zhou}]{DBLP:journals/corr/abs-2303-04048}
Jiaan Wang, Yunlong Liang, Fandong Meng, Haoxiang Shi, Zhixu Li, Jinan Xu, Jianfeng Qu, and Jie Zhou. 2023{\natexlab{a}}.
\newblock \href {https://doi.org/10.48550/arXiv.2303.04048} {Is chatgpt a good {NLG} evaluator? {A} preliminary study}.
\newblock \emph{CoRR}, abs/2303.04048.

\bibitem[{Wang et~al.(2023{\natexlab{b}})Wang, Li, Chen, Zhu, Lin, Cao, Liu, Liu, and Sui}]{DBLP:journals/corr/abs-2305-17926}
Peiyi Wang, Lei Li, Liang Chen, Dawei Zhu, Binghuai Lin, Yunbo Cao, Qi~Liu, Tianyu Liu, and Zhifang Sui. 2023{\natexlab{b}}.
\newblock \href {https://doi.org/10.48550/arXiv.2305.17926} {Large language models are not fair evaluators}.
\newblock \emph{CoRR}, abs/2305.17926.

\bibitem[{Wang et~al.(2023{\natexlab{c}})Wang, Song, Drozdov, Garimella, Manjunatha, and Iyyer}]{DBLP:journals/corr/abs-2305-14625}
Shufan Wang, Yixiao Song, Andrew Drozdov, Aparna Garimella, Varun Manjunatha, and Mohit Iyyer. 2023{\natexlab{c}}.
\newblock \href {https://doi.org/10.48550/arXiv.2305.14625} {{KNN-LM} does not improve open-ended text generation}.
\newblock \emph{CoRR}, abs/2305.14625.

\bibitem[{Yuan et~al.(2021)Yuan, Neubig, and Liu}]{DBLP:conf/nips/YuanNL21}
Weizhe Yuan, Graham Neubig, and Pengfei Liu. 2021.
\newblock \href {https://proceedings.neurips.cc/paper/2021/hash/e4d2b6e6fdeca3e60e0f1a62fee3d9dd-Abstract.html} {Bartscore: Evaluating generated text as text generation}.
\newblock In \emph{Advances in Neural Information Processing Systems 34: Annual Conference on Neural Information Processing Systems 2021, NeurIPS 2021, December 6-14, 2021, virtual}, pages 27263--27277.

\bibitem[{Zhang et~al.(2023)Zhang, Luo, Chuang, Fang, Gaitskell, Hartvigsen, Wu, Fox, Meng, and Glass}]{DBLP:journals/corr/abs-2304-03728}
Tianhua Zhang, Hongyin Luo, Yung{-}Sung Chuang, Wei Fang, Luc Gaitskell, Thomas Hartvigsen, Xixin Wu, Danny Fox, Helen Meng, and James~R. Glass. 2023.
\newblock \href {https://doi.org/10.48550/arXiv.2304.03728} {Interpretable unified language checking}.
\newblock \emph{CoRR}, abs/2304.03728.

\bibitem[{Zhang et~al.(2020)Zhang, Kishore, Wu, Weinberger, and Artzi}]{DBLP:conf/iclr/ZhangKWWA20}
Tianyi Zhang, Varsha Kishore, Felix Wu, Kilian~Q. Weinberger, and Yoav Artzi. 2020.
\newblock \href {https://openreview.net/forum?id=SkeHuCVFDr} {Bertscore: Evaluating text generation with {BERT}}.
\newblock In \emph{8th International Conference on Learning Representations, {ICLR} 2020, Addis Ababa, Ethiopia, April 26-30, 2020}. OpenReview.net.

\bibitem[{Zhao et~al.(2019)Zhao, Peyrard, Liu, Gao, Meyer, and Eger}]{zhao-etal-2019-moverscore}
Wei Zhao, Maxime Peyrard, Fei Liu, Yang Gao, Christian~M. Meyer, and Steffen Eger. 2019.
\newblock \href {https://doi.org/10.18653/v1/D19-1053} {{M}over{S}core: Text generation evaluating with contextualized embeddings and earth mover distance}.
\newblock In \emph{Proceedings of the 2019 Conference on Empirical Methods in Natural Language Processing and the 9th International Joint Conference on Natural Language Processing (EMNLP-IJCNLP)}, pages 563--578, Hong Kong, China. Association for Computational Linguistics.

\end{thebibliography}
\bibliographystyle{acl_natbib}

\appendix

\section{Examples from WikiEval}
Tables \ref{tabExamplesWikiEvalFaithfulness}, \ref{tabExamplesWikiEvalAnswerRelevance} and \ref{tabExamplesWikiEvalContextRelevance} show examples from the WikiEval dataset, focusing in particular on answers with high and low faithfulness (Table \ref{tabExamplesWikiEvalFaithfulness}), high and low answer relevance (Table \ref{tabExamplesWikiEvalAnswerRelevance}), and high and low context relevance (Table \ref{tabExamplesWikiEvalContextRelevance}).

\begin{table*}
\footnotesize
\centering
\begin{tabular}{p{90pt}p{200pt}p{120pt}}
\toprule
\textbf{Question} & \textbf{Context} & \textbf{Answer} \\
\midrule
Who directed the film Oppenheimer and who stars as J. Robert Oppenheimer in the film? &
Oppenheimer is a 2023 biographical thriller film written and directed by Christopher Nolan. Based on the 2005 biography American Prometheus by Kai Bird and Martin J. Sherwin, the film chronicles the life of J. Robert Oppenheimer, a theoretical physicist who was pivotal in developing the first nuclear weapons as part of the Manhattan Project, and thereby ushering in the Atomic Age. Cillian Murphy stars as Oppenheimer, with Emily Blunt as Oppenheimer's wife Katherine "Kitty" Oppenheimer.&
\textit{High Faithfulness}:
Christopher Nolan directed the film Oppenheimer. Cillian Murphy stars as J. Robert Oppenheimer in the film.

\smallskip
\textit{Low Faithfulness}:
James Cameron directed the film Oppenheimer. Tom Cruise stars as J. Robert Oppenheimer in the film.\\
\bottomrule
\end{tabular}
\caption{Example from WikiEval, showing answers with high and low faithfulness. \label{tabExamplesWikiEvalFaithfulness}}
\end{table*}

\begin{table*}
\footnotesize
\centering
\begin{tabular}{p{90pt}p{330pt}}
\toprule
\textbf{Question} & \textbf{Answer} \\
\midrule
When is the scheduled launch date and time for the PSLV-C56 mission, and where will it be launched from?
& \textit{High answer relevance}: The PSLV-C56 mission is scheduled to be launched on Sunday, 30 July 2023 at 06:30 IST / 01:00 UTC. It will be launched from the Satish Dhawan Space Centre, Sriharikota, Andhra Pradesh, India.

\smallskip
\textit{Low answer relevance}: The scheduled launch date and time for the PSLV-C56 mission have not been provided.The PSLV-C56 mission is an important space mission for India. It aims to launch a satellite into orbit to study weather patterns.\\
\bottomrule
\end{tabular}
\caption{Example from WikiEval, showing answers with high and low answer relevance. \label{tabExamplesWikiEvalAnswerRelevance}}
\end{table*}

\begin{table*}
\footnotesize
\centering
\begin{tabular}{p{90pt}p{330pt}}
\toprule
\textbf{Question} & \textbf{Context} \\
\midrule
When was the Chimnabai Clock Tower completed, and who was it named after?
& \textit{High context relevance}: The Chimnabai Clock Tower, also known as the Raopura Tower, is a clock tower situated in the Raopura area of Vadodara, Gujarat, India. It was completed in 1896 and named in memory of Chimnabai I (1864–1885), a queen and the first wife of Sayajirao Gaekwad III of Baroda State.

\smallskip
\textit{Low context relevance}: The Chimnabai Clock Tower, also known as the Raopura Tower, is a clock tower situated in the Raopura area of Vadodara, Gujarat, India. It was completed in 1896 and named in memory of Chimnabai I (1864–1885), a queen and the first wife of Sayajirao Gaekwad III of Baroda State. It was built in Indo-Saracenic architecture style. History. Chimnabai Clock Tower was built in 1896. The tower was named after Chimnabai I (1864–1885), a queen and the first wife of Sayajirao Gaekwad III of Baroda State. It was inaugurated by Mir Kamaluddin Hussainkhan, the last Nawab of Baroda. During the rule of Gaekwad, it was a stoppage for horse drawn trams. 
The clock tower was erected at the cost of 25,000 (equivalent to 9.2 million or USD 120,000 in 2023).\\
\bottomrule
\end{tabular}
\caption{Example from WikiEval, showing answers with high and low context relevance. \label{tabExamplesWikiEvalContextRelevance}}
\end{table*}

\end{document}